\documentclass[10pt,twocolumn,letterpaper]{article}

\usepackage[pagenumbers]{cvpr} 

\usepackage[dvipsnames]{xcolor}

\usepackage{times}
\usepackage{epsfig}
\usepackage{graphicx}
\usepackage{graphicx}
\usepackage{amsmath}
\usepackage{amssymb}

\usepackage{here}
\usepackage{caption}
\usepackage{multirow}
\usepackage{chngpage}
\usepackage{arydshln}
\usepackage{gensymb}
\usepackage{url}
\usepackage{color}
\usepackage{comment}
\usepackage{enumitem}
\usepackage{xspace}
\usepackage{multicol}

\def\eg{\emph{e.g.,}\xspace}
\def\ie{\emph{i.e.,}\xspace}

\def\etal{\emph{et al.}\xspace}

\definecolor{cvprblue}{rgb}{0.21,0.49,0.74}
\usepackage[pagebackref,breaklinks,colorlinks,citecolor=cvprblue]{hyperref}

\title{3D Human Pose Perception from Egocentric Stereo Videos}

\author{
Hiroyasu Akada\and
Jian Wang\and
Vladislav Golyanik\and
Christian Theobalt \and
Max Planck Institute for Informatics, SIC 
}
\begin{document}

\twocolumn[{%
\maketitle

\renewcommand\twocolumn[1][]{#1}%
\begin{center}
    \captionsetup{type=figure}
    \includegraphics[width=\linewidth]{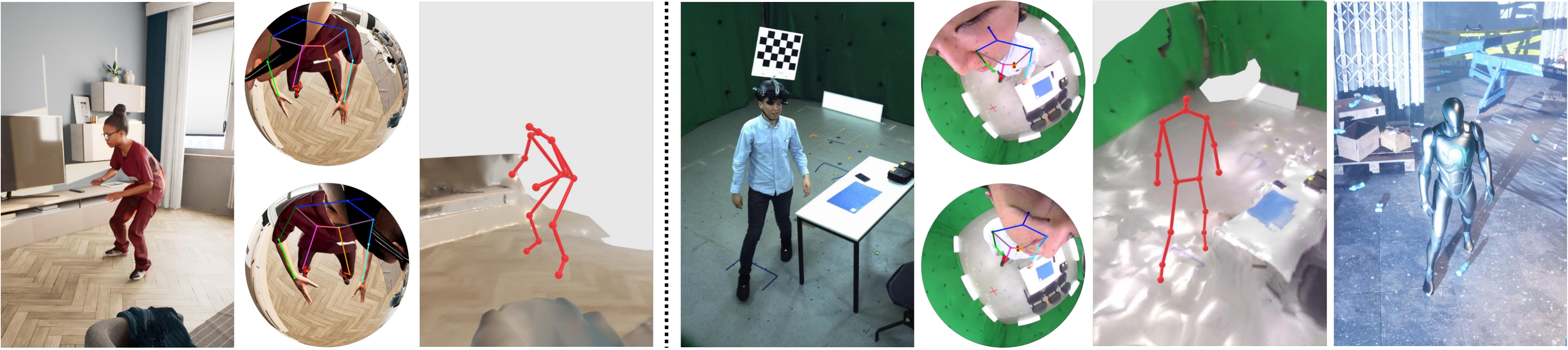}
    \vspace{-7mm}
    \caption*{\,
        (a) 
        \,\,\,\,\,\,\,\,\,\,\,\,\,\,\,\,\,\,\,\,\,\,\,\,\,\,\,\,\,\,\,\,\,\,\,
        (b) 
        \,\,\,\,\,\,\,\,\,\,\,\,\,\,\,\,\,\,\,\,\,\,\,\,\,\,\,\,\,\,\,\,\,\,
        (c) 
        \,\,\,\,\,\,\,\,\,\,\,\,\,\,\,\,\,\,\,\,\,\,\,\,\,\,\,\,\,\,\,\,\,\,\,\,\,\,\,\,\,\,\,\,\,
        (d)
        \,\,\,\,\,\,\,\,\,\,\,\,\,\,\,\,\,\,\,\,\,\,\,\,\,\,\,\,\,\,\,\,\,\,\,
        (e) 
        \,\,\,\,\,\,\,\,\,\,\,\,\,\,\,\,\,\,\,\,\,\,\,\,\,\,\,\,\,\,\,\,\,\,
        (f) 
        \,\,\,\,\,\,\,\,\,\,\,\,\,\,\,\,\,\,\,\,\,\,\,\,\,\,\,\,\,\,\,\,\,\,\,\,\,\,\,\,\,
        (g) 
    }
    \vspace{-3mm}
    \captionof{figure}{\textbf{3D human pose estimation results of our proposed method from egocentric stereo fisheye videos}. 
    \textbf{Left}: results on synthetic images; 
    (a) reference RGB view of the scene; 
    (b) 3D-to-2D pose re-projections, and (c) a 3D pose in a scene mesh reconstructed by our framework.
    \textbf{Right}: results on real-world images; (d) reference view; 
    (e) 3D-to-2D pose re-projections; (f) a 3D pose in the reconstructed scene, and (g) 3D virtual character animation (possible future application of our method). 
    }    
    \label{fig:dataset}
\end{center}%
}]

\begin{abstract}
%
%
While head-mounted devices are becoming more compact, they provide egocentric views with significant self-occlusions of the device user. 
Hence, existing methods often fail to accurately estimate complex 3D poses from egocentric views.
In this work, we propose a new transformer-based framework to improve egocentric stereo 3D human pose estimation, which leverages the scene information and temporal context of egocentric stereo videos.
Specifically, we utilize 1$)$ depth features from our 3D scene reconstruction module with uniformly sampled windows of egocentric stereo frames, and 2$)$ human joint queries enhanced by temporal features of the video inputs.
Our method is able to accurately estimate human poses even in challenging scenarios, such as crouching and sitting.
Furthermore, we introduce two new benchmark datasets, \textit{i.e.,} UnrealEgo2 and UnrealEgo-RW (RealWorld). 
%
The proposed datasets offer a much larger number of egocentric stereo views with a wider variety of human motions than the existing datasets, allowing comprehensive evaluation of existing and upcoming methods.
Our extensive experiments show that the proposed approach significantly outperforms previous methods. 
UnrealEgo2, UnrealEgo-RW, and trained models are available on our project page\footnote{\url{https://4dqv.mpi-inf.mpg.de/UnrealEgo2/}} and Benchmark Challenge\footnote{\url{https://unrealego.mpi-inf.mpg.de/}}. 


\end{abstract}    

\vspace{-4mm}

\section{Introduction} 
Egocentric 3D human motion capture using wearable devices has received increased attention recently \cite{rhodin2016egocap, xu2019mo2cap2, tome2019xr, Tom2020SelfPose3E, zhang2021automatic, wang2021estimating, wang2022estimating, jiang2021egocentric, yuan2019ego, luo2021dynamics, zhao2021egoglass, hakada2022unrealego, wang2023scene}. 
Different from traditional vision-based motion capture setups that require a fixed recording space, egocentric systems allow flexible motion capture in less constrained situations.
Therefore, the egocentric setups offer various applications, such as motion analysis and XR technologies (Fig.~\ref{fig:dataset}-(g)).

%
Previous works proposed various egocentric methods to capture device users.
On the one hand, the vast majority of existing methods---which use a monocular camera---would fail for complex human poses due to depth ambiguity and self-occlusion.
On the other hand, the methods designed for stereo devices do not yet realize the full potential of their stereo settings, especially with the most recent compact eyeglasses-based setups~\cite{zhao2021egoglass, hakada2022unrealego}.
Specifically, they do not deliver high 3D reconstruction accuracy across different scenarios. 
Moreover, these approaches do not consider scene information, which further limits their accuracy.  

%
%
%
%
%

%
To address the challenges outlined above, we propose a new transformer-based framework for egocentric 3D human motion capture from compact 
eyeglasses-based devices; 
see Fig.~\ref{fig:dataset}. 
%
%
%
%
%
The first step of our framework is to estimate 2D joint heatmaps from egocentric stereo fisheye RGB videos (Sec.~\ref{subsec:2D_module}).
%
These 2D joint heatmaps are then processed with human joint queries in our transformer-based 3D module to estimate 3D poses.
Here, we leverage the scene information and temporal context of the input videos in the 3D module to improve estimation accuracy. 
%
Firstly, we use uniformly sampled windows of egocentric stereo frames to reconstruct a 3D background scene using Structure from Motion (SfM)~\cite{schonberger2016structure}, obtaining scene depth as additional information for the 3D module (Sec.~\ref{subsec:sam_module} and~\ref{subsec:scene_module}). 
%
%
%
In our challenging eyeglasses-based setup, however, the 3D scene and camera poses can not always be estimated due to severe self-occlusion in the egocentric images.
%
%
%
This results in depth maps with zero (invalid) values and undesired computation of network gradients during training.
To mitigate this issue, we propose to use depth padding masks that prevent processing such invalid depth values in the 3D module.
Additionally, 
we propose video-dependent query augmentation that enhances the joint queries with the temporal context of stereo video inputs to effectively capture the temporal relation of human motions at a joint level (Sec.~\ref{subsec:3D_module}). 
%

%
%
%
%
%
%
%
%
%
%
%
%
%

%

We also introduce two new benchmark datasets: \textit{UnrealEgo2} and \textit{UnrealEgo-RW}.
UnrealEgo2 is an extended version of UnrealEgo~\cite{hakada2022unrealego} and the largest eyeglasses-based synthetic data with various new motions, offering $2.8 \times$ larger data ($2.5\text{M}$ images) than the existing dataset~\cite{hakada2022unrealego}. 
%
%
\text{UnrealEgo-RW} is a real-world dataset recorded with our newly developed device that resembles the virtual eyeglasses-based setup \cite{hakada2022unrealego}, offering $260\text{k}$ images with various motions and 3D poses.
The proposed datasets make it possible to evaluate existing and upcoming methods on a variety of motions, not only in synthetic scenes but also in real-world cases. 
%

%
In short, the contributions of this paper are as follows: 
\begin{itemize}
    \setlength{\itemsep}{1pt} 
    \item The transformer-based framework for egocentric stereo 3D human pose estimation that accounts for temporal context in egocentric stereo views.    
    \item 3D pose estimation is enhanced via the utilization of scene information from our video-based 3D scene reconstruction module as well as joint queries obtained from our video-dependent query augmentation policy. 
    \item A new portable device for egocentric stereo view capture with its specification and two new benchmark datasets: \textit{UnrealEgo2} and \textit{UnrealEgo-RW} recorded with our device. The proposed datasets allow for a 
    comprehensive evaluation of methods for egocentric  3D human pose estimation from stereo views. 
\end{itemize} 
Our experiments demonstrate that the proposed method outperforms the previous state-of-the-art approaches by a substantial margin, \textit{i.e.,} ${>}15\%$ on UnrealEgo~\cite{hakada2022unrealego}, ${\geq}40\%$ on UnrealEgo2, and ${\geq}10\%$ on \text{UnrealEgo-RW} (on MPJPE). 
We release UnrealEgo2, UnrealEgo-RW, and our trained models on our project page\footnote{\url{https://4dqv.mpi-inf.mpg.de/UnrealEgo2/}} and Benchmark Challenge\footnote{\url{https://unrealego.mpi-inf.mpg.de/}} to foster the area of egocentric 3D vision. 
%
%

\section{Related Work}

\noindent\textbf{Egocentric 3D Human Motion Capture.} 
Recent years witnessed significant innovations in egocentric 3D human pose estimation. 
To capture device users, many existing works use downward-facing cameras and the existing methods can be categorized into two groups. 
The first group are monocular approaches~\cite{xu2019mo2cap2, tome2019xr, Tom2020SelfPose3E, zhang2021automatic, wang2021estimating, park2023domain, wang2022estimating, jiang2021egocentric, yuan2019ego, luo2021dynamics, Liu2023, wang2024egocentric}.
For example, Wang~\etal~\cite{wang2024egocentric} uses a diffusion-based~\cite{ho2020denoising} motion prior to tackle self-conclusions.
%
%
Due to the depth ambiguity, monocular methods often fail to estimate accurate 3D poses.
Wang~\etal~\cite{wang2023scene} tackled this issue by projecting depth and 2D pose features into a pre-defined voxel space. 
This method requires additional training with ground-truth depths and human body segmentation; it cannot easily be extended for multi-view or temporal inputs. 
Zhang~\etal~\cite{Zhang_2023_ICCV} utilized a diffusion model~\cite{ho2020denoising} conditioned on a 3D scene to generate poses. 
They require pre-scanned scene mesh as an input and cannot capture a device user.

The second group, including our work, focuses on the multi-view (often stereo) setting. 
Rhodin~\etal~\cite{rhodin2016egocap} proposed an optimization approach whereas Cha~\etal\cite{8458443} used eight cameras to estimate a 3D body and reconstruct a 3D scene separately. 
Other works~\cite{zhao2021egoglass, hakada2022unrealego} used the multi-branch autoencoder~\cite{tome2019xr} to the stereo setup. 
Kang~\etal~\cite{kang2023ego3dpose} (arXiv pre-print at the time of submission) leveraged a stereo-matching mechanism and perspective embedding heatmaps.
%
In contrast to the existing methods, we propose a new transformer-based method that effectively utilizes egocentric stereo videos via our video-based 3D scene reconstruction module and video-dependent query augmentation policy.
Our method considers the scene information without the supervision of the scene data.

%
\noindent\textbf{Transformers in 3D Human Pose Estimation from External Cameras.} 
%
%
3D pose estimation from external cameras has shown significant progress due to the advances in transformer architectures~\cite{vaswani2017attention}.
%
%
Some works~\cite{lin2021end, You_2023_ICCV} predict 3D human pose and mesh from monocular views. 
Other works~\cite{pavllo:videopose3d:2019, zhu2021posegtac, Zheng_2021_ICCV, Li_2022_CVPR, li2022exploiting, zhang2022mixste, yang2022u, einfalt2023uplift, Zhao_2023_CVPR, Tang_2023_CVPR, Park_2023_ICCV, Zhu_2023_ICCV, Zhou_2023_ICCV} present a 2D-to-3D lifting module that estimates 3D poses from monocular 2D joints obtained with off-the-shelf 2D joint detectors. 
Although their lifting modules show impressive results, those monocular methods cannot be easily applied to our stereo setting. 
%
%
On the other hand, some works utilize transformers in multi-view settings. 
He~\etal~\cite{He_2020_CVPR} and Ma~\etal~\cite{ma2021transfusion} aggregate stereo information on epipolar lines of stereo images, which are difficult to obtain from fisheye images. 
Recent work~\cite{wang2021mvp} regresses multi-person 3D poses from multi-view inputs, powered by projective attention and query adaptation. 
However, no existing works explored the potential of transformers along with 2D joint heatmaps or explicit scene information in stereo 3D pose estimation.
In this paper, we propose a transformer-based framework that accounts for the temporal relation of human motion at a joint level via intermediate 2D joint heatmap and depth maps even with inaccurate depth values mixed in the framework.

\noindent\textbf{Datasets for Egocentric 3D Human Pose Estimation.} 
Several works proposed unique setups to create datasets, using a monocular camera~\cite{xu2019mo2cap2, tome2019xr, wang2021estimating, wang2022estimating, jiang2021egocentric, yuan2019ego, luo2021dynamics, Li2023EgoBody} and forward-facing cameras~\cite{jiang2021egocentric, yuan2019ego, luo2021dynamics, Zhang_ECCV_2022, Pan_2023_ICCV, Khirodkar_2023_ICCV, Zhang_2023_ICCV, Li2023EgoBody}.
%
There also exist datasets captured with stereo devices~\cite{rhodin2016egocap, zhao2021egoglass, 8458443, Grauman_etal_2022_CVPR, Pan_2023_ICCV, Khirodkar_2023_ICCV}.
However, they are small~\cite{rhodin2016egocap} with limited motion types~\cite{rhodin2016egocap, zhao2021egoglass}, not publicly available~\cite{zhao2021egoglass, 8458443}, or do not provide ground truth 3D poses of device users~\cite{Grauman_etal_2022_CVPR, Pan_2023_ICCV, Khirodkar_2023_ICCV}.
%
%
Recently, Akada~\etal~\cite{hakada2022unrealego} introduced UnrealEgo, a synthetic dataset based on virtual eyeglasses with two fisheye cameras. 
However, they provide only synthetic images.
%
%
Meanwhile, more glasses-based stereo datasets that offer a wider variety of motions or real-world footage are required nowadays for an extensive evaluation of existing and upcoming methods. 
Hence, we introduce two new benchmark datasets that in their characteristics go beyond the existing data: \textit{UnrealEgo2} and \textit{UnrealEgo-RW}. 
%
%
We describe the proposed datasets in the following section.

\section{Mobile Device and Datasets}
\label{sec:datasets} 
%
We present two new datasets for egocentric stereo 3D motion capture: UnrealEgo2 and UnrealEgo-RW; see Fig.~\ref{fig:dataset}. 
%
%
Please watch our supplementary video for visualizations.
\noindent\textbf{UnrealEgo2 Dataset.} 
To create UnrealEgo2 (an extension of UnrealEgo~\cite{hakada2022unrealego}), we adapt the publicly available setup with a virtual eyeglasses device~\cite{hakada2022unrealego}. 
This setup comes with two downward-facing fisheye cameras attached $12\mathrm{cm}$ apart from each other on the glasses frames. 
The camera's field of view is $170\degree$. 
With this device, we capture $17$ realistic 3D human models~\cite{renderpeople} animated by the Mixamo~\cite{mixamo} dataset in various 3D environments. 
We record simple to highly complex motions such as crouching and crawling, for 14 hours. 

Overall, UnrealEgo2 offers 15,207 motions and 
${>}1.25 \text{M}$ stereo views ($2.5 \text{M}$ images) as well as depth maps with a resolution $1024{\times}1024$ pixel rendered at $25$ frames per second. 
Each frame is annotated with $32$ body and $40$ hand joints. 
Note that UnrealEgo2 is the largest glasses-based dataset and $2.8 \times$ larger than UnrealEgo. 
Also, it does not share the same motions with  UnrealEgo, providing a larger motion variety for a 
comprehensive evaluation. 

%

\begin{figure}[t]
 \centering
 \includegraphics[width=\linewidth]{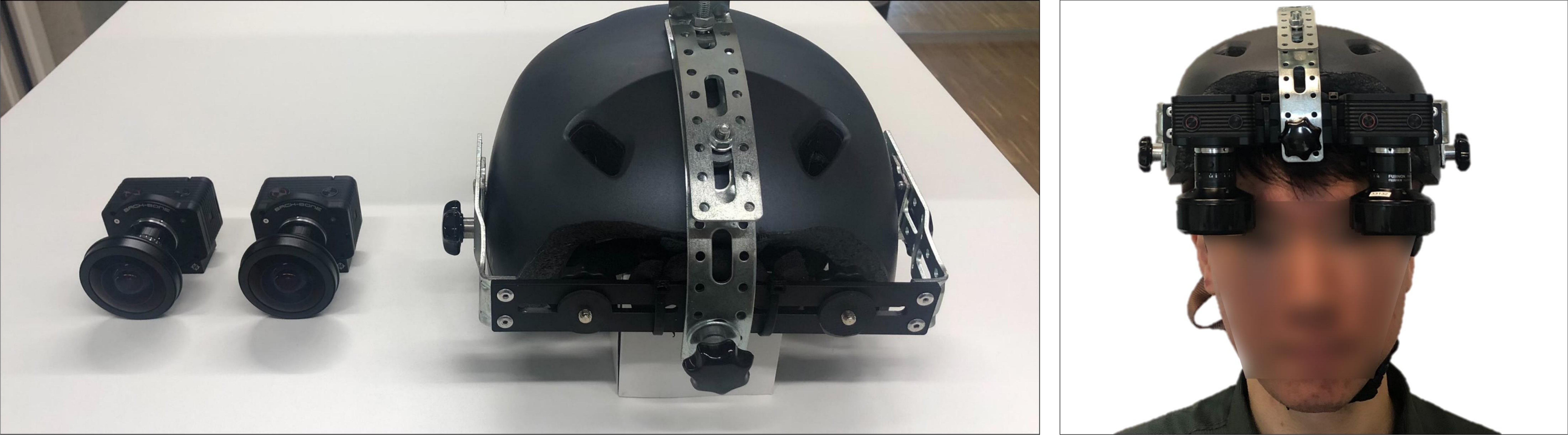}
 \vspace{-6mm}
 \caption{Our portable setup to acquire \textit{UnrealEgo-RW}.} 
 \label{fig:device} 
\end{figure}

%
\noindent\textbf{Design of Our Mobile Device.} 
Evaluation with real-world datasets plays a pivotal role in computer vision research. 
Therefore, we build a new portable device; see Fig.~\ref{fig:device}. 
Our device is based on a helmet with two RIBCAGE RX0 \text{I\hspace{-1.2pt}I} cameras~\cite{ribcagecamera} and two FUJINON FE185C057HA-1 fisheye lenses~\cite{fujinonlens}. 
We placed the cameras 12cm away from each other and 2cm away from user's face.
We cropped the margins of the egocentric images to resemble the field of view of $170\degree$ of the UnrealEgo and UnrealEgo2 setups. 
Note that our setup is more compact than EgoCap~\cite{rhodin2016egocap} that placed cameras 25cm away from user's face.

\noindent\textbf{UnrealEgo-RW (Real-World) Dataset.} 
With our device, we record various motions of 16 identities in a multi-view motion capture studio (Fig.~\ref{fig:dataset}-(d)). 
%
%
We capture simple and challenging activities, \eg crawling and dancing, for 1.5 hours. 
This is in strong contrast to the existing real-world stereo dataset~\cite{zhao2021egoglass} (not publicly available) that records only three simple actions, \ie sitting, standing, and walking.
%
%
%
%
%
%
%
%
%

%
%
%
%

In total, we obtained 591 motion segments from 16 identities with various textured clothing. 
This results in more than $130\text{k}$ stereo views ($260\text{k}$ images) of a resolution $872{\times}872$ pixel rendered at $25$ frames per second with ground-truth 3D poses of 16 joints.
Note that UnrealEgo-RW offers $4.3 \times$ larger data with a wider variety of motions than the publicly available real-world stereo data~\cite{rhodin2016egocap}.


%

%
%
%

\begin{figure*}[t]
 \centering
  \includegraphics[width=\linewidth]{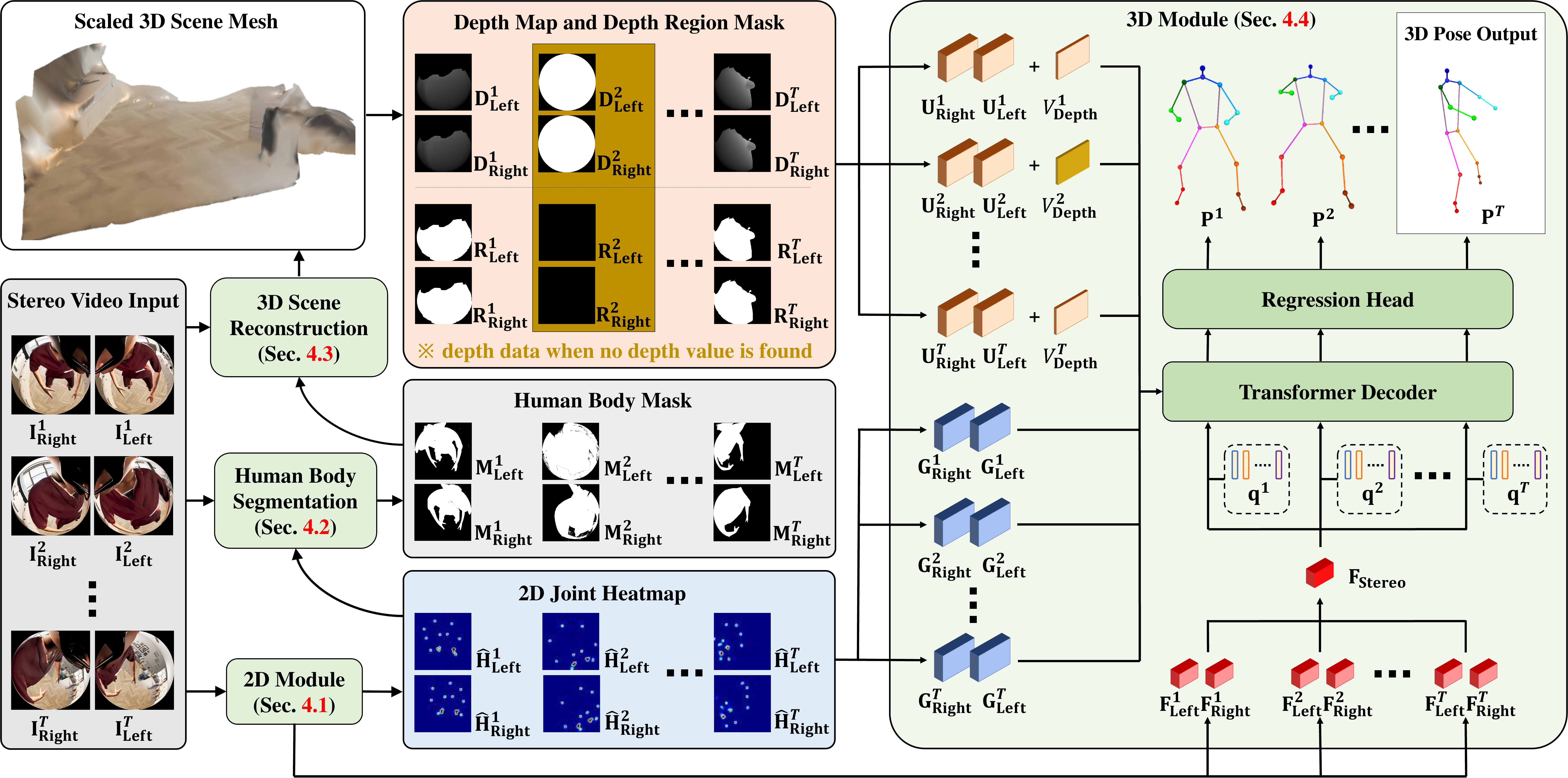}
 \vspace{-6mm}
 \caption{
 \textbf{Overview of our framework.} 
 Our method takes egocentric stereo videos $\{\mathbf{I}^{t}_{\text{Left}}, \mathbf{I}^{t}_{\text{Right}}\}$ as inputs. 
 We first apply the 2D module to obtain 2D joint heatmaps $\{\mathbf{H}^{t}_{\text{Left}}, \mathbf{H}^{t}_{\text{Right}}\}$ and video features $\{\mathbf{F}^{t}_{\text{Left}}, \mathbf{F}^{t}_{\text{Right}}\}$ (Sec.~\ref{subsec:2D_module}). 
 The heatmaps are used with input videos to create human body masks $\{{\mathbf{M}}^{t}_{\text{Left}}, {\mathbf{M}}^{t}_{\text{Right}} \}$ (Sec.~\ref{subsec:sam_module}). 
 Next, we use uniformly sampled windows of input frames and human body masks to reconstruct a 3D scene mesh (Sec.~\ref{subsec:scene_module}). 
 From the mesh, we generate depth maps $\{{\mathbf{D}}^{t}_{\text{Left}}, {\mathbf{D}}^{t}_{\text{Right}} \}$ and depth region masks $\{{\mathbf{R}}^{t}_{\text{Left}}, {\mathbf{R}}^{t}_{\text{Right}}\}$. 
 Note that this diagram shows an example case of missing depth values for the second input frame.
 Lastly, the depth data, 2D joint heatmaps, video features, joint queries $q^{t}$ and the padding masks ${V}^{t}_{\text{Depth}}$ are processed in the 3D module to estimate 3D poses ${\mathbf{P}}^{t}$ (Sec.~\ref{subsec:3D_module}).
}
 \label{fig:overview method}
\end{figure*}

\section{Method} 

We propose a new framework for egocentric stereo 3D human pose estimation as shown in Fig.~\ref{fig:overview method}.
Our framework first estimates the 2D joint heatmaps from egocentric stereo fisheye videos in our 2D module (Sec.~\ref{subsec:2D_module}).
The heatmaps and input videos are then processed in our segmentation module to obtain 2D human body masks (Sec.~\ref{subsec:sam_module}).
Next, we use uniformly sampled windows of input frames and human body masks to reconstruct 3D scenes (Sec.~\ref{subsec:scene_module}).
Here, we render depth maps and depth region masks from the reconstructed mesh.
Finally, our transformer-based 3D module processes the joint heatmaps, depth information, and joint queries to estimate 3D poses (Sec.~\ref{subsec:3D_module}).
Here, the 3D module leverages depth padding masks based on the availability of the depth maps as well as joint queries enhanced by the stereo video features from the 2D module.

\subsection{2D Pose Estimation} 
\label{subsec:2D_module}

Given egocentric stereo videos with $T$ frames $\{\mathbf{I}^{t}_{\text{Left}}, \mathbf{I}^{t}_{\text{Right}} \in \mathbb{R}^{H \times W \times 3 } | t = 1, 2, \ldots, T \}$, we use the existing stereo 2D joint heatmap estimator~\cite{hakada2022unrealego} to obtain a sequence of corresponding 2D heatmaps of 15 joints $\{\mathbf{H}^{t}_{\text{Left}}, \mathbf{H}^{t}_{\text{Right}} \in \mathbb{R}^{\frac{H}{4} \times \frac{W}{4} \times 15 }\}$, including the neck, upper arms, lower arms, hands, thighs, calves, feet, and balls of the feet.
%
%
%
%
We also extract intermediate feature maps $\{\mathbf{F}^{t}_{\text{Left}}, \mathbf{F}^{t}_{\text{Right}} \in \mathbb{R}^{\frac{H}{32} \times \frac{W}{32} \times C } \}$ where $C = 512$, which are used later in the 3D module.
%



\subsection{Human Body Segmentation} 
\label{subsec:sam_module}

To reconstruct 3D scenes from egocentric videos, it is necessary to identify the pixels corresponding to the background environment.
Therefore, we integrate an existing segmentation method, \ie ViT-H SAM model~\cite{kirillov2023segany}, as our segmentation network $\mathcal{F}_{\text{SAM}}$.
In this module, we firstly obtain 2D joint locations from the 2D joint heatmap $\{\widehat{\mathbf{H}}^{t}_{\text{Left}}, \widehat{\mathbf{H}}^{t}_{\text{Right}}\}$. 
Then, we use the input video frames $\{\mathbf{I}^{t}_{\text{Left}}, \mathbf{I}^{t}_{\text{Right}}\}$ and its corresponding 2D joints to extract a human body mask $\{{\mathbf{M}}^{t}_{\text{Left}}, {\mathbf{M}}^{t}_{\text{Right}}  \in \mathbb{R}^{H \times W \times 1 } \}$:
%
\begin{equation}
  {\mathbf{M}}^{t}_{\text{Left}} = \mathcal{F}_{\text{SAM}}(\mathbf{I}^{t}_{\text{Left}}, \widehat{\mathbf{H}}^{t}_{\text{Left}} ).
\end{equation}
The same process can be applied to obtain ${\mathbf{M}}^{t}_{\text{Right}}$.
Note that we use the SAM model without re-training on ground-truth human body masks. Instead, we guide the predictions of SAM using joint positions extracted from the 2D heatmaps.

\subsection{3D Scene Reconstruction} 
\label{subsec:scene_module}

%
We aim to reconstruct 3D environments from uniformly sampled windows of input frames $\{\mathbf{I}^{t}_{\text{Left}}, \mathbf{I}^{t}_{\text{Right}}\}$ and human body masks $\{{\mathbf{M}}^{t}_{\text{Left}}, {\mathbf{M}}^{t}_{\text{Right}}\}$ with a fixed length.
The length is set to 4 seconds (some motion data contains shorter sequences). 
Given these data, we use Metashape~\cite{metashape} to perform SfM to obtain camera poses and a 3D scene mesh.
%
Here, as the baseline length between stereo cameras is known, \ie 12cm, we can obtain the mesh in the real-world scale.
%
Next, we render down-sampled depth maps $\{{\mathbf{D}}^{t}_{\text{Left}}, {\mathbf{D}}^{t}_{\text{Right}}  \in \mathbb{R}^{\frac{H}{4} \times \frac{W}{4} \times 1 } \}$ and depth region masks $\{{\mathbf{R}}^{t}_{\text{Left}}, {\mathbf{R}}^{t}_{\text{Right}}  \in \mathbb{R}^{\frac{H}{4} \times \frac{W}{4} \times 1} \}$ from the reconstructed 3D scene mesh. 
The depth region masks show the regions where the depth values are obtained from the 3D scene.
%
This depth information will be used later in the 3D module as additional cues for pose estimation.
However, there are some cases where the egocentric RGB videos are largely occupied by a human body. 
In such scenarios, the 3D scene can not be reconstructed or camera poses can not be estimated.
This results in missing (invalid) depth values and undesired computation of network gradients during training.
Therefore, we tackle this issue in our 3D module.
%

\subsection{3D Pose Estimation} 
\label{subsec:3D_module}
In the 3D module, we aim to estimate a sequence of 3D poses by considering scene information and the temporal context of the egocentric stereo videos.
%
Specifically, given the 2D joint heatmaps, depth maps, depth region masks, and $T$ sets of joint queries $q^{t} \in \mathbb{R}^{16 \times \frac{C}{2} }$, we use a transformer decoder to estimate a sequence of 3D poses $\{ {{\mathbf{P}}^{t} \in \mathbb{R}^{16\times3} | t = 1, 2, \ldots, T \} }$. 
Our pose output is the 3D pose at the last time step ${\mathbf{P}}^{T}$.
We follow the existing works~\cite{tome2019xr, Tom2020SelfPose3E, hakada2022unrealego} to estimate 16 joints including the head.

\begin{table}[t]
\begin{center}
\scalebox{0.675}{
\begin{tabular}{llcccc}
\noalign{\smallskip}
\hline
\noalign{\smallskip}
\multicolumn{1}{l}{Method} & \multicolumn{1}{c}{Task} & \multicolumn{1}{c}{MPJPE$(\downarrow)$} & \multicolumn{1}{c}{PA-MPJPE$(\downarrow)$} & \multicolumn{1}{c}{3D PCA$(\uparrow)$} & \multicolumn{1}{c}{AUC$(\uparrow)$}\\ 
\noalign{\smallskip}
\hline
\noalign{\smallskip}
Zhao~\etal~\cite{zhao2021egoglass} & \multirow{5}{*}{\begin{tabular}[l]{@{}c@{}}Pelvis \\ relative \end{tabular}} & 86.45 & 63.71 & 85.97 & 50.50 \\
Akada~\etal~\cite{hakada2022unrealego}  &  & 78.98 & 59.30 & 88.81 & 54.31 \\ 
Kang~\etal~\cite{kang2023ego3dpose}  &  & 60.82 & \underline{48.47} & - & - \\
Baseline &  & \underline{59.85}  & 49.14 & \underline{92.07} & \underline{63.88} \\
Ours  &  & \bf 50.55  & \bf 40.50 & \bf 93.83 & \bf 70.61 \\
\noalign{\smallskip}
\cdashline{1-6}
\noalign{\smallskip}
Zhao~\etal~\cite{zhao2021egoglass} & \multirow{4}{*}{\begin{tabular}[l]{@{}c@{}}Device \\ relative \end{tabular}} & 88.12 & 65.36 & 85.10 & 50.37 \\
Akada~\etal~\cite{hakada2022unrealego} &  & 84.53 & 63.92 & 87.05 & 52.76 \\ 
Baseline &  & \underline{63.44} & \underline{50.97}  & \underline{92.30} & \underline{64.54}\\
Ours &  & \bf 46.20 & \bf 40.19 & \bf 94.02 & \bf 73.53 \\
\noalign{\smallskip}
\hline
\noalign{\smallskip}
\end{tabular}
}
\end{center}
\vspace{-6mm}
\caption{Quantitative results on UnrealEgo~\cite{hakada2022unrealego} with mm-scale.  
}
\label{table:quantitative_unrealego}
\end{table}

\noindent\textbf{Depth and Heatmap Features.}
We use the sequence of the depth maps, depth region masks, and the 2D joint heatmaps as the memory of a cross-attention operation in the transformer decoder.
For this purpose, we extract depth features $\{{\mathbf{U}}^{t}_{\text{Left}}, {\mathbf{U}}^{t}_{\text{Right}}  \in \mathbb{R}^{\frac{H}{32} \times \frac{W}{32} \times \frac{C}{2}} \}$ from the depth data:
\begin{equation}
  {\mathbf{U}}^{t}_{\text{Left}} = \mathcal{F}_{\text{Depth}}(\mathbf{D}^{t}_{\text{Left}} \oplus \widehat{\mathbf{R}}^{t}_{\text{Left}} ), 
\end{equation}
where ``$\oplus$'' is a concatenation operation along the channel axis and $\mathcal{F}_{\text{Depth}}$ represents a feature extractor. The same process can be applied to obtain ${\mathbf{U}}^{t}_{\text{Right}}$.

Similarly, we extract heatmap features $\{{\mathbf{G}}^{t}_{\text{Left}}, {\mathbf{G}}^{t}_{\text{Right}}  \in \mathbb{R}^{\frac{H}{16} \times \frac{W}{16} \times C} \}$ from the 2D heatmaps:
\begin{equation}
  {\mathbf{G}}^{t}_{\text{Left}} =  \mathcal{F}_{\text{HM}}(\widehat{\mathbf{H}}^{t}_{\text{Left}}), 
\end{equation}
where $\mathcal{F}_{\text{HM}}$ represents another feature extractor. The same process can be applied to obtain ${\mathbf{G}}^{t}_{\text{Right}}$.
%

These features are forwarded with positional embeddings into the transformer.
%
%
%
However, as mentioned in Sec.~\ref{subsec:scene_module}, depth values can be missing in some frames.
To prevent processing features of such depth data and let the network focus only on valid frames, we propose to add padding masks ${V}^{t}_{\text{Depth}}\!\in \mathcal{R}$ to all the elements of $\{{\mathbf{U}}^{t}_{\text{Left}}, {\mathbf{U}}^{t}_{\text{Right}} \}$: 
%
\begin{equation}
    V^{t}_{\text{Depth}} = 
        \begin{cases}
            \text{-inf},& \text{if depth values are missing}\\
            0, & \text{otherwise}
        \end{cases}.
\end{equation}
%
%
When ${V}^{t}_{\text{Depth}} = \text{-inf}$, the depth features $\{{\mathbf{U}}^{t}_{\text{Left}}, {\mathbf{U}}^{t}_{\text{Right}} \}$ after the softmax function in self-attention layers of the transformer will have zero effect on the network training.
%

\begin{table}[t]
\begin{center}
\scalebox{0.765}{
\begin{tabular}{lcccc}
\noalign{\smallskip}
\hline
\noalign{\smallskip}
\multicolumn{1}{l}{Method} & \multicolumn{1}{c}{MPJPE$(\downarrow)$} & \multicolumn{1}{c}{PA-MPJPE$(\downarrow)$} & \multicolumn{1}{c}{3D PCA$(\uparrow)$} & \multicolumn{1}{c}{AUC$(\uparrow)$}\\ 
\noalign{\smallskip}
\hline
\noalign{\smallskip}
Zhao~\etal~\cite{zhao2021egoglass} & 79.64 & 58.22 & 88.50 & 53.82 \\
Akada~\etal~\cite{hakada2022unrealego} & 72.80 & 52.88 & 91.32 & 55.81 \\ 
Baseline & \underline{52.23} & \underline{39.78}  & \underline{95.72} & \underline{68.13}\\
Ours & \bf 30.53 & \bf 26.72 & \bf 97.22 & \bf 80.75 \\
\noalign{\smallskip}
\hline
\noalign{\smallskip}
\end{tabular}
}
\end{center}
\vspace{-6mm}
\caption{Quantitative results of device-relative pose estimation on UnrealEgo2 with mm-scale.  
}
\label{table:quantitative_unrealego2}
\end{table}

\noindent\textbf{Stereo-Video-Dependent Joint Query Adaptation.} 
The existing work~\cite{wang2021mvp} represents human joints as learnable positional embeddings called joint queries that encode prior knowledge about the skeleton joints.
In our problem setting, the simplest way to design such joint queries is to set queries for each pose in a motion sequence.
However, this can not capture the temporal context in video inputs, \eg human motions and background changes.
Therefore, we extend the multi-view joint query augmentation technique~\cite{wang2021mvp} for our stereo video setting to account for sequential information.
Specifically, we enhance the joint queries with the temporal intermediate features of stereo RGB frames $\{ \mathbf{F}^{t}_{\text{Left}}, \mathbf{F}^{t}_{\text{Right}} \}$.
Firstly, from the sequence of the intermediate features, we create a sequence of combined features $\mathbf{F}^{t}  \in \mathbb{R}^{\frac{H}{32} \times \frac{W}{32} \times \frac{C}{2} }$:
\begin{equation}
  \mathbf{F}^{t}  = \operatorname{conv}(\mathbf{F}^{t}_{\text{Left}} \oplus \mathbf{F}^{t}_{\text{right}}),
\end{equation}
where ``$\operatorname{conv}(\cdot)$'' is a convolution operation with a kernel size of $1 \times 1$.

Next, we fuse the sequence of the combined features $\mathbf{F}^{t}$ to obtain a fused stereo features $\mathbf{F}_{\text{Stereo}} \in \mathbb{R}^{\frac{C}{2}}$:
%
\begin{equation}
\mathbf{F}_{\text{Stereo}}  = \mathbf{F}^{1}_{\text{P}} \oplus \ldots \oplus\mathbf{F}^{T}_{\text{P}} ,  \, \text{where} \,\, \mathbf{F}^{i}_{\text{P}} = p(\mathbf{F}^{i}),
\end{equation}
where ``$p(\cdot)$'' is an operation of adaptive average pooling. Now, the feature $\mathbf{F}_{\text{Stereo}}$ contains stereo video information.

Lastly, with $\mathbf{F}_{\text{Stereo}}$ and a fully connected layer $"\operatorname{fc}(\cdot)"$, we augment each query $q^{t}$ to obtain $q^{t}_{\text{Aug}} \in \mathbb{R}^{16 \times \frac{C}{2}}$:
\begin{equation}
  {\mathbf{q}}^{t}_{\text{Aug}} = \text{fc}(\mathbf{F}_{\text{Stereo}}) + q^{t}.
\end{equation}
%

\begin{table}[t]
\begin{center}
\scalebox{0.765}{
\begin{tabular}{lcccc}
\noalign{\smallskip}
\hline
\noalign{\smallskip}
\multicolumn{1}{l}{Method} & \multicolumn{1}{c}{MPJPE$(\downarrow)$} & \multicolumn{1}{c}{PA-MPJPE$(\downarrow)$} & \multicolumn{1}{c}{3D PCA$(\uparrow)$} & \multicolumn{1}{c}{AUC$(\uparrow)$}\\ 
\noalign{\smallskip}
\hline
\noalign{\smallskip}
Zhao~\etal~\cite{zhao2021egoglass} & 117.57 & 88.01 & 73.12 & 38.94 \\
Akada~\etal~\cite{hakada2022unrealego} & 122.64 & 86.55 & 72.51 & 38.67 \\ 
Baseline & \underline{115.95} & \underline{85.00}  & \underline{74.13} & \underline{40.11}\\
Ours & \bf 104.14 & \bf 82.18 & \bf 80.20 & \bf 46.22 \\
\noalign{\smallskip}
\hline
\noalign{\smallskip}
\end{tabular}
}
\end{center}
\vspace{-6mm}
\caption{Quantitative results of device-relative pose estimation on UnrealEgo-RW with mm-scale.  
}
\label{table:quantitative_unrealego_rw}
\vspace{-2mm}
\end{table}

%
\begin{figure*}[t]
 \centering
 \includegraphics[width=\linewidth]{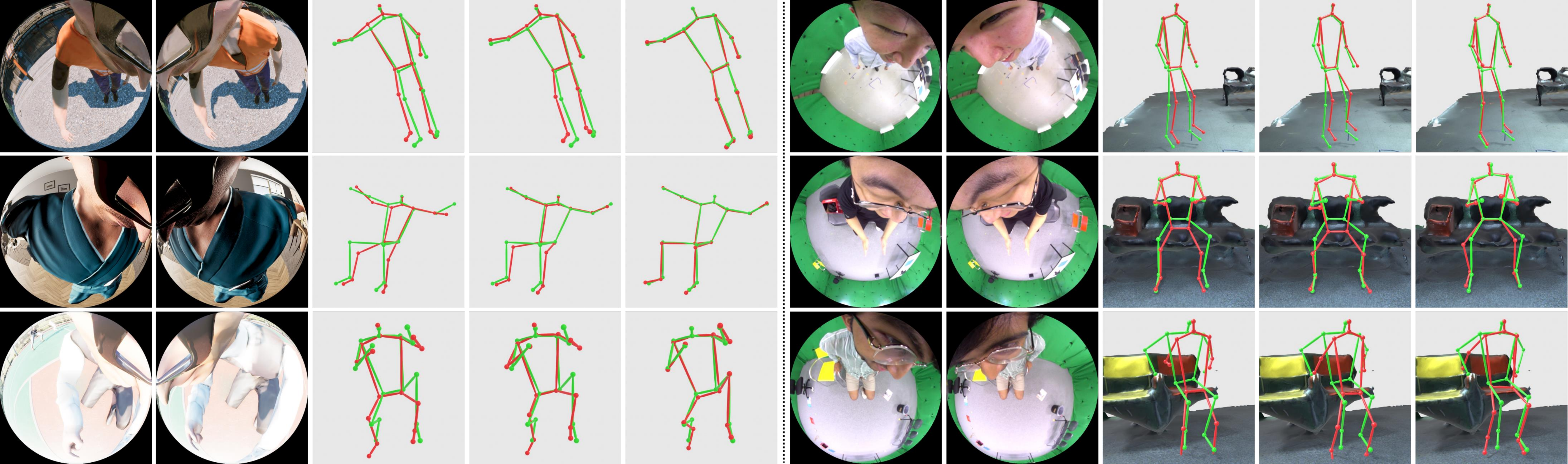}
     \vspace{-7mm}
    \caption*{\footnotesize
        \,\,\,\,\,\,\,\,\,\,\,\,
        Stereo inputs
        \,\,\,\,\,\,\,\,\,\,\,\,\,\,\,\,\,\,
        Akada~\etal~\cite{hakada2022unrealego}
        \,\,\,\,\,\,
        Baseline        
        \,\,\,\,\,\,\,\,\,\,\,\,\,\,\,\,\,\,
        Ours
        \,\,\,\,\,\,\,\,\,\,\,\,\,\,\,\,\,\,\,\,\,\,\,\,\,\,\,\,\,\,\,\,\,\,
        Stereo inputs
        \,\,\,\,\,\,\,\,\,\,\,\,\,\,\,\,\,\,
        Akada~\etal~\cite{hakada2022unrealego}
        \,\,\,\,
        Baseline        
        \,\,\,\,\,\,\,\,\,\,\,\,\,\,\,\,\,
        Ours
        \,\,\,\,
    }
    \vspace{-2mm}
 \caption{Qualitative results of device-relative pose estimation. \textbf{Left}: UnrealEgo2. \textbf{Right}: UnrealEgo-RW. 3D pose prediction and ground truth are displayed in red and green, respectively. For UnrealEgo-RW, we show ground-truth scene meshes for visualization. 
 }
 \label{fig:qualitaitve_result}
\end{figure*}

\noindent\textbf{Transformer Decoder.}
We adopt a DETR~\cite{carion2020end}-based transformer decoder and a pose regression head.
In decoder layers, all of the augmented joint queries $q^{t}_{\text{Aug}}$ first interact with each other on a self-attention layer.
Then, the queries extract all of the temporal stereo features from the memory $\{{\mathbf{U}}^{t}_{\text{Left}}, {\mathbf{U}}^{t}_{\text{Right}}, {\mathbf{G}}^{t}_{\text{Left}}, {\mathbf{G}}^{t}_{\text{Right}}\}$ with the padding masks ${V}^{t}_{\text{Depth}}$ on a cross-attention layer.
Lastly, the pose regression head estimates a sequence of 3D poses $\{ \hat{\mathbf{P}^{t}} \in \mathbb{R}^{16\times3}  | t = 1, 2, \ldots, T \}$, yielding the final pose output ${\mathbf{P}}^{T}$.

Similar to the previous works~\cite{zhang2022mixste, einfalt2023uplift}, we train the 3D module with the pose supervision of the current and past frames:
\begin{equation}
 \begin{split}
  L_{\text{3D}} = L_{\text{pose}}(\mathbf{P}^{T}, \hat{\mathbf{P}}^{T}) + \frac{\lambda_{\text{past}}}{(T-1)}\sum_{t=1}^{T-1}L_{\text{pose}}(\mathbf{P}^{t}, \hat{\mathbf{P}}^{t}),
  \end{split}
\end{equation}
%
%
%
\begin{equation} 
 \begin{split}
  L_{\text{pose}}(\mathbf{P}, \hat{\mathbf{P}}) = & \lambda_{\text{pose}}(\operatorname{mpjpe}(\mathbf{P}, \hat{\mathbf{P}}) + \\
  & \lambda_{\operatorname{cos}}\operatorname{cos}(\operatorname{bone}(\mathbf{P}), \operatorname{bone}(\hat{\mathbf{P}}))),
  \end{split}
\end{equation}
where $\mathbf{P}$ is a ground-truth 3D pose, mpjpe($\cdot$) is the mean per joint position error, cos($\cdot$) is a negative cosine similarity, and bone($\cdot$) is an operation of obtaining bones of the 3D poses as used in the previous work~\cite{hakada2022unrealego}:
\begin{equation}
  \operatorname{mpjpe}(\mathbf{P}, \hat{\mathbf{P}}) = \frac{1}{NJ}\sum_{n=1}^{N}\sum_{j=1}^{J}||\mathbf{P}_{n, j} - \hat{\mathbf{P}}_{n,j}||_2, 
\end{equation}
\begin{equation}
  \operatorname{cos}(\mathbf{B}, \hat{\mathbf{B}}) = -\frac{1}{N}\sum_{n=1}^{N}\sum_{m=1}^{M}\frac{\mathbf{B}_{n,m} \cdot \hat{\mathbf{B}}_{n,m} }{||\mathbf{B}_{n,m}|| \, ||\hat{\mathbf{B}}_{n,m}||},
\end{equation} 
where $N$ is batch size, $J$ is the number of joints, $M$ is the number of bones, and $\mathbf{B}_{n,m} \in \mathbb{R}^3$ is a vector of $m$-th bone. 
%
%

\begin{figure*}[t]
 \centering
 \includegraphics[width=1.00\linewidth]{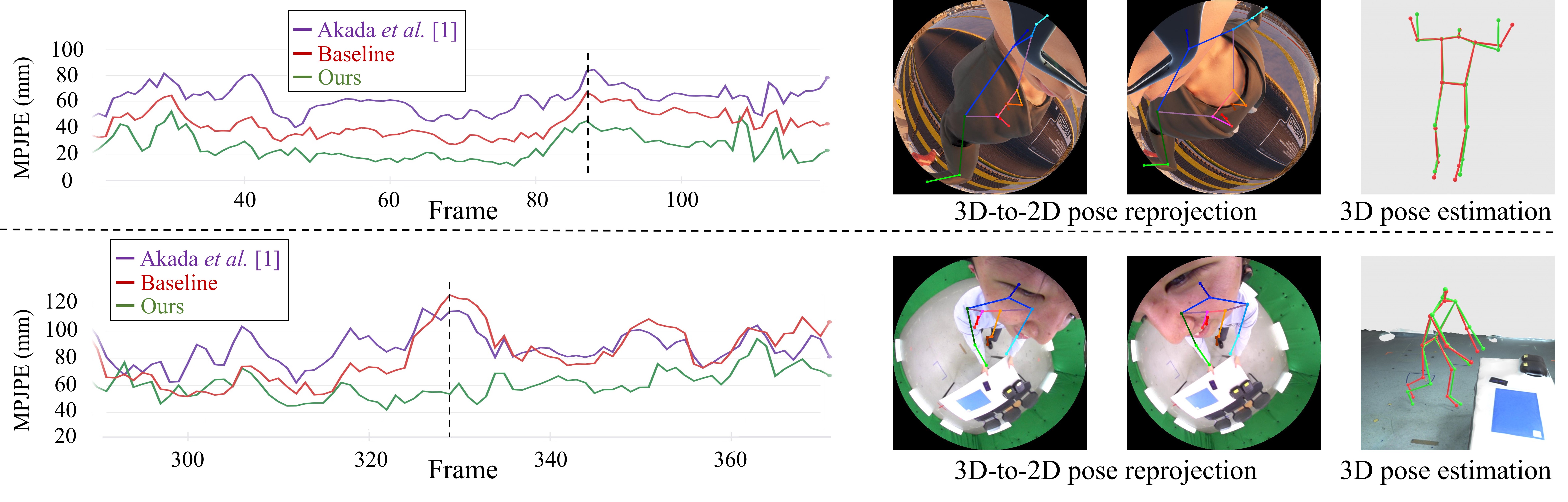}
     \vspace{-8mm}
    \caption{Results of our framework and comparison methods on example sequences from UnrealEgo2 (above) and UnrealEgo-RW (below). \textbf{Left}: MPJPE curves. \textbf{Right}: Outputs of our method at frame 87 and 329 of the sequences, respectively. 3D pose estimation and ground truth are colored in red and green, respectively.
    }
 \label{fig:qualitaitve_result_MPJPE}
\vspace{-3mm}
\end{figure*}


\begin{table}[t]
\begin{center}
\scalebox{0.76}{
\begin{tabular}{lccc}
\noalign{\smallskip}
\hline
\noalign{\smallskip}
\multicolumn{1}{l}{Method} & \multicolumn{1}{c}{MPJPE$(\downarrow)$} & \multicolumn{1}{c}{PA-MPJPE$(\downarrow)$} \\ 

\noalign{\smallskip}
\hline
\noalign{\smallskip}

(a) Baseline with depth information & 120.39 & 86.23 \\
Baseline & \bf 115.36 & \bf 84.80 \\

\noalign{\smallskip}
\hline
\noalign{\smallskip}

(b) Ours w/o query adaptation & 108.33 & 86.69 \\
(c) Ours w/o depth information & 112.56 & 84.37 \\
(d) Ours w/o depth padding mask & 108.70 & 84.26 \\

\noalign{\smallskip}
\cdashline{1-4}
\noalign{\smallskip}

(e) Ours with latest pose supervision only & 105.67 & 83.46 \\
(f) Ours with a single set of queries & 105.58 & 85.68 \\

\noalign{\smallskip}
\cdashline{1-4}
\noalign{\smallskip}

Ours & \bf 104.14 & \bf 82.18 \\

\noalign{\smallskip}
\hline
\noalign{\smallskip}
\end{tabular}
}
\end{center}
\vspace{-6mm}
\caption{Ablation study of our model for device-relative pose estimation on UnrealEgo-RW with mm-scale.}
\label{table:ablation_method}
\end{table}

\section{Experiments}

\subsection{Datasets for Evaluation}
We use three datasets for our experiments: UnrealEgo~\cite{hakada2022unrealego}, UnrealEgo2, and UnrealEgo-RW.
For UnrealEgo, we use their proposed data splits.
Also, we divide UnrealEgo2 into 12,139 motions (1,002,656 stereo views) for training, 1,545 motions (127,968 stereo views) for validation, and 1523 motions (123,488 stereo views) for testing. 
Similarly, we split UnrealEgo-RW into 547 motions (51,936 stereo views) for training, 77 motions (7,616 stereo views) for validation, and 86 motions (7,936 stereo views) for testing.
we follow the existing works~\cite{xu2019mo2cap2, tome2019xr, Tom2020SelfPose3E, zhang2021automatic, wang2021estimating, wang2022estimating, jiang2021egocentric, zhao2021egoglass, hakada2022unrealego, wang2023scene} to report the results of device-relative 3D pose estimation.
For UnrealEgo, we also follow the existing works~\cite{hakada2022unrealego, kang2023ego3dpose} to include the results of pelvis-relative 3D pose estimation.


\subsection{Training Details}
\label{subsec:training_details} 
%
We resize the input RGB images and ground-truth 2D keypoint heatmaps to $256{\times}256$ and $64{\times}64$ pixels, respectively. 
For the training of the 2D module, we follow the previous work~\cite{hakada2022unrealego} to use the ResNet18~\cite{he2016deep} pre-traned on ImageNet~\cite{imagenet_cvpr09} as an encoder and train the module with a batch size of 16 and an initial learning rate of $10^{-3}$. 
Then, we train the 3D module with a batch size of 32 and an initial learning rate of $2 \cdot 10^{-4}$. 
The modules are trained with Adam optimizer~\cite{Kingma2015} for ten epochs, starting with the initial learning rate for the first half epochs and applying a linearly decaying rate for the next half. 
Also, we set the hyper-parameters as $\lambda_{\text{pose}}=0.1$ $\lambda_{\text{cos}}=0.01$, and $\lambda_{\text{past}}=0.1$. 
We use five sequential stereo views as inputs to our model, \ie $T=5$, with a skip size of 3. 
See our supplement for more details on the network architecture.

\begin{table}[t]
\begin{center}
\scalebox{0.655}{
\begin{tabular}{lcccc}
\noalign{\smallskip}
\hline
\noalign{\smallskip}
\multicolumn{1}{l}{Method} & \multicolumn{1}{c}{\begin{tabular}[l]{@{}c@{}}Upper body \\ MPJPE$(\downarrow)$ \end{tabular}} & \multicolumn{1}{c}{\begin{tabular}[l]{@{}c@{}}Lower body \\ MPJPE$(\downarrow)$ \end{tabular}}  & \multicolumn{1}{c}{\begin{tabular}[l]{@{}c@{}}Foot \\ MPJPE$(\downarrow)$ \end{tabular}} & \multicolumn{1}{c}{\begin{tabular}[l]{@{}c@{}}Foot \\ MPE$(\downarrow)$ \end{tabular}}  \\ 
\noalign{\smallskip}
\hline
\noalign{\smallskip}
Ours w/o depth information & 80.82 & 144.31 & 174.45 & 6.39 \\
Ours w/o depth padding masks & \bf 77.29 & \underline{140.10} & \underline{169.95} & \underline{5.02} \\
Ours & \underline{77.85} & \bf 130.97 & \bf 155.86 & \bf{4.83}\\

\noalign{\smallskip}
\hline
\noalign{\smallskip}
\end{tabular}
}
\end{center}
\vspace{-6mm}
\caption{The effect of scene information (depth) per body part on UnrealEgo-RW. The numbers are in $mm$.} 
\label{table:ablation_depth}
\end{table}

\subsection{Evaluation}
We compare our method with existing stereo-based egocentric pose estimation methods~\cite{zhao2021egoglass, hakada2022unrealego}. 
%
%
We use the official source code of Akada~\etal~\cite{hakada2022unrealego} and re-implement the framework of Zhao~\etal~\cite{zhao2021egoglass} as its source code is not available.
Note that the comparison methods are trained on the same datasets as our model.
%
%
Kang~\etal~\cite{kang2023ego3dpose} (arXiv pre-print at the time of submission) only shows results of the pelvis-relative estimation on UnrealEgo. 
Therefore, we include them for reference.
Furthermore, we are interested in the performance of the publicly available state-of-the-art method~\cite{hakada2022unrealego} with temporal inputs. 
Thus, we modify their 3D module such that it can take as an input a sequence of stereo 2D keypoint heatmaps with the same time step as ours, \ie $T = 5$. 
Here, we replace the first and the last fully connected layers in the encoder, the pose decoder, and the heatmap reconstruction decoder of their autoencoder-based 3D module~\cite{hakada2022unrealego} by those with $T$ times the size of the original hidden dimension.
We denote this model as Baseline and train it with the same training procedure as Akada~\etal~\cite{hakada2022unrealego}. 
%
%
Note that Akada~\etal~\cite{hakada2022unrealego}, Baseline, and our model use the same 2D module.
%

We follow the existing works~\cite{xu2019mo2cap2, tome2019xr, Tom2020SelfPose3E, zhang2021automatic, wang2021estimating, wang2022estimating, jiang2021egocentric, zhao2021egoglass, hakada2022unrealego, wang2023scene} to report Mean Per Joint Position Error (MPJPE) and Mean Per Joint Position Error with Procrustes Alignment~\cite{10.1214/ss/1177012582} (PA-MPJPE). 
%
We additionally report 3D Percentage of Correct Keypoints (3D PCK) and Area Under the Curve (AUC) for UnrealEgo2 and UnrealEgo-RW. 
%

%

\noindent\textbf{Results on Synthetic Datasets.} Tables~\ref{table:quantitative_unrealego} and~\ref{table:quantitative_unrealego2} report the results with UnrealEgo~\cite{hakada2022unrealego} and UnrealEgo2.
Our method outperforms the existing methods~\cite{zhao2021egoglass, hakada2022unrealego, kang2023ego3dpose} and Baseline across all metrics by a significant margin, \eg ${>}15\%$ on UnrealEgo~\cite{hakada2022unrealego} and ${\geq}40\%$ on UnrealEgo2 (on MPJPE). 
%
The qualitative results on UnrealEgo2 in Fig.~\ref{fig:qualitaitve_result}-(left part)
show that existing methods and Baseline fail to estimate lower bodies of complex poses with severe self-occlusions, such as crouching. 
Even under such challenging scenarios, however, our approach yields accurate 3D poses.
See Fig.~\ref{fig:qualitaitve_result_MPJPE}-(above part) 
for a MPJPE curve and visual outputs of our framework on UnrealEgo2.
Our method is able to constantly estimate accurate 3D poses compared to the existing methods.
As evidenced by these results, our method demonstrates superiority and becomes a strong benchmark method in the egocentric stereo 3D pose estimation tasks.
See our supplementary material and video for more results.

\begin{table}[t]
\begin{center}
\scalebox{0.765}{
\begin{tabular}{lcccc}
\noalign{\smallskip}
\hline
\noalign{\smallskip}
\multicolumn{1}{l}{Method} & \multicolumn{1}{c}{MPJPE$(\downarrow)$} & \multicolumn{1}{c}{PA-MPJPE$(\downarrow)$} & \multicolumn{1}{c}{3D PCA$(\uparrow)$} & \multicolumn{1}{c}{AUC$(\uparrow)$}\\ 
\noalign{\smallskip}
\hline
\noalign{\smallskip}
T = 1 & 108.63 & 84.69 & 77.98 & 44.15 \\
T = 3 & 108.23 & 85.28 & 78.35 & 44.54 \\
T = 5 & \underline{104.14} & \bf 82.18 & \underline{80.20} & \bf 46.22 \\
T = 7 & \bf 104.01 & \underline{82.43} & \bf 80.52 & \underline{46.10} \\
\noalign{\smallskip}
\hline
\noalign{\smallskip}
\end{tabular}
}
\end{center}
\vspace{-6mm}
\caption{Ablation study of our model with different sequence lengths on UnrealEgo-RW. The numbers are in $mm$.} 
\label{table:ablation_sequence}
\vspace{-5mm}
\end{table}

\begin{figure*}[t]
 \centering
 \includegraphics[width=\linewidth]{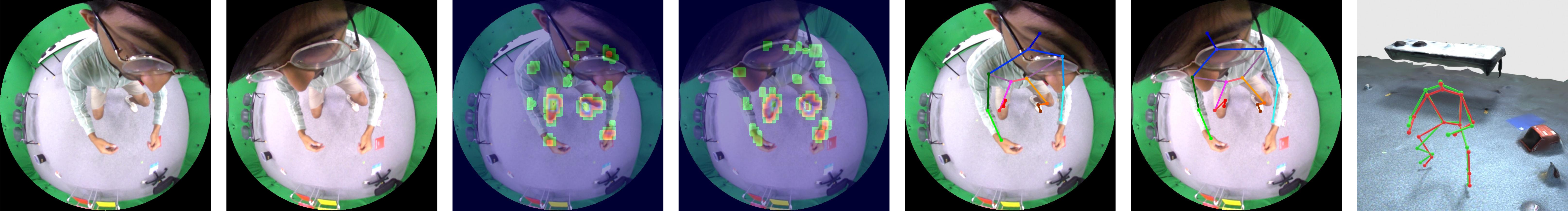}
     \vspace{-7mm}
    \caption*{\footnotesize
        \,\,\,\,\,\,\,\,\,\,\,\,\,\,\,\,\,\,\,\,\,\,\,\,\,\,\,\,\,\,\,\,\,\,\,
        Stereo inputs
        \,\,\,\,\,\,\,\,\,\,\,\,\,\,\,\,\,\,\,\,\,\,\,\,\,\,\,
        \,\,\,\,\,\,\,\,\,\,\,\,\,\,\,\,\,\,\,\,\,\,\,\,\,\,\,
        2D joint heatmap estimation
        \,\,\,\,\,\,\,\,\,\,\,\,\,\,\,\,\,\,
        \,\,\,\,\,\,\,\,\,\,\,\,\,\,\,\,\,\,
        3D-to-2D pose reprojection        
        \,\,\,\,\,\,\,\,\,\,\,\,\,\,\,\,\,\,\,\,
        3D pose estimation
        \,
    }
    \vspace{-2mm} 
 \caption{Visualization of outputs from our model on UnrealEgo-RW. 3D-to-2D pose reprojection is visualized in the same colors as in Fig.~\ref{fig:dataset}-(e). 3D pose estimation and ground truth are displayed in red and green, respectively.
 }
 \label{fig:qualitaitve_result_ours}
 \vspace{-2mm}
\end{figure*}

\noindent\textbf{Results on the Real-World Dataset.}
Table~\ref{table:quantitative_unrealego_rw} shows quantitative results on UnrealEgo-RW.
Again, our method outperforms the existing methods~\cite{zhao2021egoglass, hakada2022unrealego} and Baseline across all metrics, \eg by more than 10\% on MPJPE.
See Fig.~\ref{fig:qualitaitve_result}-(right) for qualitative results. 
The current state-of-the-art methods~\cite{zhao2021egoglass, hakada2022unrealego} or Baseline show floating feet, inaccurate pelvis position, and penetration to the floor ground.
However, our method is able to estimate accurate 3D poses.
See Fig.~\ref{fig:qualitaitve_result_MPJPE}-(below part) for a MPJPE curve and visual outputs on an example motion of UnrealEgo-RW.
The curve indicates that our method constantly shows lower 3D errors than the comparison methods.
All of the results indicate the effectiveness of our proposed framework compared to the existing methods.
We also visualize 2D heatmaps, 3D-to-2D pose reprojection, and 3D pose prediction from our method in Fig.~\ref{fig:qualitaitve_result_ours}.
Even when the joint locations of the lower body are estimated closely in the 2D heatmaps, our approach predicts accurate lower body poses.
%
%
These results suggest that the proposed method with our portable device can open up the possibility of many future applications, including animating virtual humans (Fig.~\ref{fig:dataset}-(g)). 
For the virtual human animation, we applied inverse kinematics with estimated 3D joint locations and ground-truth camera poses to drive the character in a world coordinate system.


\noindent\textbf{Ablation Study.}
In Table~\ref{table:ablation_method}, we first ablate (a) the CNN-based 3D module (Baseline) with depth data concatenated to the heatmap inputs.
However, naively adding this extra scene information to this 3D module does not help probably because the CNN layers can be affected by invalid depth values even with the depth region masks. 
%
%

Next, we test our transformer-based 3D module (b) without query augmentation (c) without depth data.
They perform worse than our full framework. 
%
%
We also ablate our method (d) without the padding mask.
The result indicates that adding depth padding masks helps because the padding mask can filter out the invalid values in depth maps in the attention module.
%
%
These results validate that our video-based 3D scene reconstruction module and video-dependent query augmentation policy boost 3D joint localization accuracy. 
Next, we ablate our model (e) with 3D pose supervision of the latest frame only. 
Note that this ablation uses the same sets of input data and joint queries as the original model, \ie $T=5$.
This model estimates less accurate poses due to the loss of supervision from past 3D poses.
We also test (f) a single set of joint queries, \ie $q^{1}$, instead of $T$ sets to predict the latest 3D pose.
Similar to (e), this model cannot benefit from the supervision of past 3D poses.

We further investigate the effect of the scene information.  Table~\ref{table:ablation_depth} shows the MPJPE per body part and Mean Penetration Error (MPE)~\cite{PhysCapTOG2020, PhysAwareTOG2021} between feet and floor ground. 
The results reveal that depth features with the padding masks reduce the errors in the lower body while maintaining the performance in the upper body.

In Table~\ref{table:ablation_sequence}, we ablate the effect of the sequence length of input frames for our method.
It is worth noting that our model with $T{=}1$ yields better results than the best existing method~\cite{hakada2022unrealego} and Baseline that utilizes temporal information (see Table~\ref{table:quantitative_unrealego_rw}).
Since our model uses the same 2D module as Akada~\etal~\cite{hakada2022unrealego} and Baseline, the difference comes only from the 3D module.
This suggests that their autoencoder-based 3D modules with the heatmap reconstruction component are, very likely, not the most suitable solution 
for estimating 3D poses from 2D joint heatmaps, highlighting the potential of our transformer-based framework. 
The result also indicates that although the longer sequence can bring performance improvement to some extent, the sequence lengths of five and seven show comparable results.

\noindent\textbf{Synthetic Data for Pre-training.}
No existing works explored the efficacy of synthetic data for pre-training in egocentric 3D pose estimation.
%
%
Thus, we further conduct experiments with models pre-trained on the synthetic datasets and fine-tuned on the real-world data.
%
Tables~\ref{table:quantitative_unrealego_rw} and~\ref{table:fine_turning} show that all methods benefit from the training with the large-scale synthetic data even with the differences in the synthetic and real-world setups, \eg fisheye distortion and syn-to-real domain gaps.
Note that the gain of our method from UnrealEgo to UnrealEgo2 is significant, \ie 3.3\% on MPJPE (75.34mm to 72.89mm). 
%
%
This suggests that it is helpful to develop not only new models but also large-scale synthetic datasets even with different distortion and domain gaps.
%
%

%
\begin{table}[t]
\begin{center}
\scalebox{0.625}{
\begin{tabular}{llcccc}
\noalign{\smallskip}
\hline
\noalign{\smallskip}
\multicolumn{1}{l}{Method} & \multicolumn{1}{c}{\begin{tabular}[l]{@{}c@{}}Initial \\ training data\end{tabular}} & \multicolumn{1}{c}{MPJPE$(\downarrow)$} & \multicolumn{1}{c}{PA-MPJPE$(\downarrow)$} & \multicolumn{1}{c}{3D PCA$(\uparrow)$} & \multicolumn{1}{c}{AUC$(\uparrow)$}\\ 
\noalign{\smallskip}
\hline
\noalign{\smallskip}
Zhao~\etal~\cite{zhao2021egoglass} & \multirow{4}{*}{UnrealEgo~\cite{hakada2022unrealego}} & 99.09 & 72.47 & 79.82 & 43.55  \\
Akada~\etal~\cite{hakada2022unrealego}  &  & 94.87 & 69.79 & 82.78 & 46.80 \\ 
Baseline &  & \underline{83.89}  & \underline{64.30} & \underline{86.20} & \underline{51.63}\\
Ours  &  & \bf75.34  & \bf 57.29 & \bf 89.43 & \bf 55.77 \\
\noalign{\smallskip}
\cdashline{1-6}
\noalign{\smallskip}
Zhao~\etal~\cite{zhao2021egoglass} & \multirow{4}{*}{UnrealEgo2} & 97.86 & 69.92 & 81.53 & 46.32 \\
Akada~\etal~\cite{hakada2022unrealego} &  & 92.48 & 67.15 & 84.25 & 48.04 \\ 
Baseline &  & \underline{82.16} & \underline{61.60}  & \underline{87.07} & \underline{52.72}\\
Ours &  & \bf 72.89 & \bf 56.19 & \bf 90.29 & \bf 57.19 \\
\noalign{\smallskip}
\hline
\noalign{\smallskip}
\end{tabular}
}
\end{center}
\vspace{-6mm}
\caption{Fine-tuning results of device-relative 3D pose estimation on UnrealEgo-RW with mm-scale. 
}
\label{table:fine_turning}
\vspace{-2mm}
\end{table}

\section{Conclusion}
In this paper, we proposed a new transformer-based framework that significantly boosts the accuracy of egocentric stereo 3D human pose estimation. 
The proposed framework leverages the scene information and temporal context of egocentric stereo video inputs via our video-based 3D scene reconstruction module and video-based joint query augmentation policy.
Our extensive experiments on the new synthetic and real-world datasets with challenging human motions validate the effectiveness of our approach compared to the existing methods.
We hope that our proposed benchmark datasets and trained models will foster the further development of methods for egocentric 3D vision. 

\noindent\textbf{Acknowledgment.}
The work was supported by the ERC Consolidator Grant 4DReply (770784) and the Nakajima Foundation. We thank Silicon Studio Corp. for providing the fisheye plug-in for Unreal Engine.


{
    \small
    \bibliographystyle{ieeenat_fullname}
    \bibliography{main}

\begin{thebibliography}{58}
\providecommand{\natexlab}[1]{#1}
\providecommand{\url}[1]{\texttt{#1}}
\expandafter\ifx\csname urlstyle\endcsname\relax
  \providecommand{\doi}[1]{doi: #1}\else
  \providecommand{\doi}{doi: \begingroup \urlstyle{rm}\Url}\fi

\bibitem[Akada et~al.(2022)Akada, Wang, Shimada, Takahashi, Theobalt, and Golyanik]{hakada2022unrealego}
Hiroyasu Akada, Jian Wang, Soshi Shimada, Masaki Takahashi, Christian Theobalt, and Vladislav Golyanik.
\newblock Unrealego: A new dataset for robust egocentric 3d human motion capture.
\newblock In \emph{European Conference on Computer Vision (ECCV)}, 2022.

\bibitem[Carion et~al.(2020)Carion, Massa, Synnaeve, Usunier, Kirillov, and Zagoruyko]{carion2020end}
Nicolas Carion, Francisco Massa, Gabriel Synnaeve, Nicolas Usunier, Alexander Kirillov, and Sergey Zagoruyko.
\newblock End-to-end object detection with transformers.
\newblock In \emph{European Conference on Computer Vision (ECCV)}, 2020.

\bibitem[Cha et~al.(2018)Cha, Price, Wei, Lu, Rewkowski, Chabra, Qin, Kim, Su, Liu, Ilie, State, Xu, Frahm, and Fuchs]{8458443}
Young-Woon Cha, True Price, Zhen Wei, Xinran Lu, Nicholas Rewkowski, Rohan Chabra, Zihe Qin, Hyounghun Kim, Zhaoqi Su, Yebin Liu, Adrian Ilie, Andrei State, Zhenlin Xu, Jan-Michael Frahm, and Henry Fuchs.
\newblock Towards fully mobile 3d face, body, and environment capture using only head-worn cameras.
\newblock \emph{IEEE Transactions on Visualization and Computer Graphics}, 24\penalty0 (11):\penalty0 2993--3004, 2018.

\bibitem[Deng et~al.(2009)Deng, Dong, Socher, Li, Li, and Fei-Fei]{imagenet_cvpr09}
J. Deng, W. Dong, R. Socher, L.-J. Li, K. Li, and L. Fei-Fei.
\newblock {ImageNet: A Large-Scale Hierarchical Image Database}.
\newblock In \emph{Computer Vision and Pattern Recognition (CVPR)}, 2009.

\bibitem[Einfalt et~al.(2023)Einfalt, Ludwig, and Lienhart]{einfalt2023uplift}
Moritz Einfalt, Katja Ludwig, and Rainer Lienhart.
\newblock Uplift and upsample: Efficient 3d human pose estimation with uplifting transformers.
\newblock In \emph{Winter Conference on Applications of Computer Vision (WACV)}, 2023.

\bibitem[FUJINON FE185C057HA-1 fisheye lens()]{fujinonlens}
FUJINON FE185C057HA-1 fisheye lens, 2023.
\newblock \url{https://www.fujifilm.com/de/de/business/optical-devices/mvlens/fe185}.

\bibitem[Grauman et~al.(2022)Grauman, Westbury, Byrne, Chavis, Furnari, Girdhar, Hamburger, et~al.]{Grauman_etal_2022_CVPR}
Kristen Grauman, Andrew Westbury, Eugene Byrne, Zachary Chavis, Antonino Furnari, Rohit Girdhar, Jackson Hamburger, et~al.
\newblock Ego4d: Around the world in 3,000 hours of egocentric video.
\newblock In \emph{Computer Vision and Pattern Recognition (CVPR)}, 2022.

\bibitem[He et~al.(2016)He, Zhang, Ren, and Sun]{he2016deep}
Kaiming He, Xiangyu Zhang, Shaoqing Ren, and Jian Sun.
\newblock Deep residual learning for image recognition.
\newblock In \emph{Computer Vision and Pattern Recognition (CVPR)}, 2016.

\bibitem[He et~al.(2020)He, Yan, Fragkiadaki, and Yu]{He_2020_CVPR}
Yihui He, Rui Yan, Katerina Fragkiadaki, and Shoou-I Yu.
\newblock Epipolar transformers.
\newblock In \emph{Computer Vision and Pattern Recognition (CVPR)}, 2020.

\bibitem[Ho et~al.(2020)Ho, Jain, and Abbeel]{ho2020denoising}
Jonathan Ho, Ajay Jain, and Pieter Abbeel.
\newblock Denoising diffusion probabilistic models.
\newblock 2020.

\bibitem[Jiang and Ithapu(2021)]{jiang2021egocentric}
Hao Jiang and Vamsi~Krishna Ithapu.
\newblock Egocentric pose estimation from human vision span.
\newblock In \emph{International Conference on Computer Vision (ICCV)}, 2021.

\bibitem[Kang et~al.(2023)Kang, Lee, Zhang, and Lee]{kang2023ego3dpose}
Taeho Kang, Kyungjin Lee, Jinrui Zhang, and Youngki Lee.
\newblock Ego3dpose: Capturing 3d cues from binocular egocentric views.
\newblock In \emph{SIGGRAPH Asia Conference}, 2023.

\bibitem[Kendall(1989)]{10.1214/ss/1177012582}
David~G. Kendall.
\newblock {A Survey of the Statistical Theory of Shape}.
\newblock \emph{Statistical Science}, 4\penalty0 (2):\penalty0 87 -- 99, 1989.

\bibitem[Khirodkar et~al.(2023)Khirodkar, Bansal, Ma, Newcombe, Vo, and Kitani]{Khirodkar_2023_ICCV}
Rawal Khirodkar, Aayush Bansal, Lingni Ma, Richard Newcombe, Minh Vo, and Kris Kitani.
\newblock Ego-humans: An ego-centric 3d multi-human benchmark.
\newblock In \emph{International Conference on Computer Vision (ICCV)}, 2023.

\bibitem[Kingma and Ba(2015)]{Kingma2015}
Diederik Kingma and Jimmy Ba.
\newblock Adam: A method for stochastic optimization.
\newblock In \emph{International Conference on Learning Representations (ICLR)}, 2015.

\bibitem[Kirillov et~al.(2023)Kirillov, Mintun, Ravi, Mao, Rolland, Gustafson, Xiao, Whitehead, Berg, Lo, Doll{\'a}r, and Girshick]{kirillov2023segany}
Alexander Kirillov, Eric Mintun, Nikhila Ravi, Hanzi Mao, Chloe Rolland, Laura Gustafson, Tete Xiao, Spencer Whitehead, Alexander~C. Berg, Wan-Yen Lo, Piotr Doll{\'a}r, and Ross Girshick.
\newblock Segment anything.
\newblock \emph{arXiv:2304.02643}, 2023.

\bibitem[Li et~al.(2023)Li, Liu, and Wu]{Li2023EgoBody}
Jiaman Li, Karen Liu, and Jiajun Wu.
\newblock Ego-body pose estimation via ego-head pose estimation.
\newblock In \emph{Computer Vision and Pattern Recognition (CVPR)}, 2023.

\bibitem[Li et~al.(2022{\natexlab{a}})Li, Liu, Ding, Liu, Wang, and Yang]{li2022exploiting}
Wenhao Li, Hong Liu, Runwei Ding, Mengyuan Liu, Pichao Wang, and Wenming Yang.
\newblock Exploiting temporal contexts with strided transformer for 3d human pose estimation.
\newblock \emph{IEEE Transactions on Multimedia (TMM)}, 2022{\natexlab{a}}.

\bibitem[Li et~al.(2022{\natexlab{b}})Li, Liu, Tang, Wang, and Van~Gool]{Li_2022_CVPR}
Wenhao Li, Hong Liu, Hao Tang, Pichao Wang, and Luc Van~Gool.
\newblock Mhformer: Multi-hypothesis transformer for 3d human pose estimation.
\newblock In \emph{Computer Vision and Pattern Recognition (CVPR)}, 2022{\natexlab{b}}.

\bibitem[Lin et~al.(2021)Lin, Wang, and Liu]{lin2021end}
Kevin Lin, Lijuan Wang, and Zicheng Liu.
\newblock End-to-end human pose and mesh reconstruction with transformers.
\newblock In \emph{Computer Vision and Pattern Recognition (CVPR)}, 2021.

\bibitem[Liu et~al.(2023)Liu, Yang, Gu, Chen, Guo, and Yang]{Liu2023}
Yuxuan Liu, Jianxin Yang, Xiao Gu, Yijun Chen, Yao Guo, and Guang-Zhong Yang.
\newblock Egofish3d: Egocentric 3d pose estimation from a fisheye camera via self-supervised learning.
\newblock \emph{IEEE Transactions on Multimedia (TMM)}, pages 1--12, 2023.

\bibitem[Luo et~al.(2021)Luo, Hachiuma, Yuan, and Kitani]{luo2021dynamics}
Zhengyi Luo, Ryo Hachiuma, Ye Yuan, and Kris Kitani.
\newblock Dynamics-regulated kinematic policy for egocentric pose estimation.
\newblock 2021.

\bibitem[Ma et~al.(2021)Ma, Chen, Kong, Wang, Liu, Tang, Yan, Xie, Lin, and Xie]{ma2021transfusion}
Haoyu Ma, Liangjian Chen, Deying Kong, Zhe Wang, Xingwei Liu, Hao Tang, Xiangyi Yan, Yusheng Xie, Shih-Yao Lin, and Xiaohui Xie.
\newblock Transfusion: Cross-view fusion with transformer for 3d human pose estimation.
\newblock In \emph{British Machine Vision Conference (BMVC)}, 2021.

\bibitem[Metashape()]{metashape}
Metashape, 2023.
\newblock \url{https://www.agisoft.com/}.

\bibitem[Mixamo()]{mixamo}
Mixamo, 2022.
\newblock \url{https://www.mixamo.com}.

\bibitem[Pan et~al.(2023)Pan, Charron, Yang, Peters, Whelan, Kong, Parkhi, Newcombe, and Ren]{Pan_2023_ICCV}
Xiaqing Pan, Nicholas Charron, Yongqian Yang, Scott Peters, Thomas Whelan, Chen Kong, Omkar Parkhi, Richard Newcombe, and Yuheng~(Carl) Ren.
\newblock Aria digital twin: A new benchmark dataset for egocentric 3d machine perception.
\newblock In \emph{International Conference on Computer Vision (ICCV)}, 2023.

\bibitem[Park et~al.(2023{\natexlab{a}})Park, Kaai, Hossain, Sumi, Rambhatla, and Fieguth]{park2023domain}
Jinman Park, Kimathi Kaai, Saad Hossain, Norikatsu Sumi, Sirisha Rambhatla, and Paul Fieguth.
\newblock Domain-guided spatio-temporal self-attention for egocentric 3d pose estimation.
\newblock In \emph{Conference on Knowledge Discovery and Data Mining (KDD)}, 2023{\natexlab{a}}.

\bibitem[Park et~al.(2023{\natexlab{b}})Park, You, Lee, and Lee]{Park_2023_ICCV}
Sungchan Park, Eunyi You, Inhoe Lee, and Joonseok Lee.
\newblock Towards robust and smooth 3d multi-person pose estimation from monocular videos in the wild.
\newblock In \emph{International Conference on Computer Vision (ICCV)}, 2023{\natexlab{b}}.

\bibitem[Pavllo et~al.(2019)Pavllo, Feichtenhofer, Grangier, and Auli]{pavllo:videopose3d:2019}
Dario Pavllo, Christoph Feichtenhofer, David Grangier, and Michael Auli.
\newblock 3d human pose estimation in video with temporal convolutions and semi-supervised training.
\newblock In \emph{Computer Vision and Pattern Recognition (CVPR)}, 2019.

\bibitem[RenderPeople()]{renderpeople}
RenderPeople, 2022.
\newblock \url{https://renderpeople.com}.

\bibitem[Rhodin et~al.(2016)Rhodin, Richardt, Casas, Insafutdinov, Shafiei, Seidel, Schiele, and Theobalt]{rhodin2016egocap}
Helge Rhodin, Christian Richardt, Dan Casas, Eldar Insafutdinov, Mohammad Shafiei, Hans-Peter Seidel, Bernt Schiele, and Christian Theobalt.
\newblock Egocap: egocentric marker-less motion capture with two fisheye cameras.
\newblock \emph{ACM Transactions on Graphics (TOG)}, 35\penalty0 (6):\penalty0 1--11, 2016.

\bibitem[RIBCAGE RX0 \text{I\hspace{-1.2pt}I} camera()]{ribcagecamera}
RIBCAGE RX0 \text{I\hspace{-1.2pt}I} camera, 2023.
\newblock \url{https://www.back-bone.ca/product/ribcage-rx0-2/}.

\bibitem[Schonberger and Frahm(2016)]{schonberger2016structure}
Johannes~L Schonberger and Jan-Michael Frahm.
\newblock Structure-from-motion revisited.
\newblock In \emph{Computer Vision and Pattern Recognition (CVPR)}, 2016.

\bibitem[Shimada et~al.(2020)Shimada, Golyanik, Xu, and Theobalt]{PhysCapTOG2020}
Soshi Shimada, Vladislav Golyanik, Weipeng Xu, and Christian Theobalt.
\newblock Physcap: Physically plausible monocular 3d motion capture in real time.
\newblock \emph{ACM Transactions on Graphics (TOG)}, 39\penalty0 (6), 2020.

\bibitem[Shimada et~al.(2021)Shimada, Golyanik, Xu, P\'{e}rez, and Theobalt]{PhysAwareTOG2021}
Soshi Shimada, Vladislav Golyanik, Weipeng Xu, Patrick P\'{e}rez, and Christian Theobalt.
\newblock Neural monocular 3d human motion capture with physical awareness.
\newblock \emph{ACM Transactions on Graphics (TOG)}, 40\penalty0 (4), 2021.

\bibitem[Tang et~al.(2023)Tang, Qiu, Hao, Hong, and Yao]{Tang_2023_CVPR}
Zhenhua Tang, Zhaofan Qiu, Yanbin Hao, Richang Hong, and Ting Yao.
\newblock 3d human pose estimation with spatio-temporal criss-cross attention.
\newblock In \emph{Conference on Computer Vision and Pattern Recognition (CVPR)}, 2023.

\bibitem[Tome et~al.(2019)Tome, Peluse, Agapito, and Badino]{tome2019xr}
Denis Tome, Patrick Peluse, Lourdes Agapito, and Hernan Badino.
\newblock xr-egopose: Egocentric 3d human pose from an hmd camera.
\newblock In \emph{International Conference on Computer Vision (ICCV)}, 2019.

\bibitem[Tome et~al.(2023)Tome, Alldieck, Peluse, Pons-Moll, Agapito, Badino, and de~la Torre]{Tom2020SelfPose3E}
Denis Tome, Thiemo Alldieck, Patrick Peluse, Gerard Pons-Moll, Lourdes Agapito, Hernan Badino, and Fernando de~la Torre.
\newblock Selfpose: 3d egocentric pose estimation from a headset mounted camera.
\newblock \emph{IEEE Transactions on Pattern Analysis and Machine Intelligence (PAMI)}, 45\penalty0 (6):\penalty0 6794--6806, 2023.

\bibitem[Vaswani et~al.(2017)Vaswani, Shazeer, Parmar, Uszkoreit, Jones, Gomez, Kaiser, and Polosukhin]{vaswani2017attention}
Ashish Vaswani, Noam Shazeer, Niki Parmar, Jakob Uszkoreit, Llion Jones, Aidan~N Gomez, {\L}ukasz Kaiser, and Illia Polosukhin.
\newblock Attention is all you need.
\newblock In \emph{Advances in neural information processing systems (NeurIPS)}, 2017.

\bibitem[Wang et~al.(2021{\natexlab{a}})Wang, Liu, Xu, Sarkar, and Theobalt]{wang2021estimating}
Jian Wang, Lingjie Liu, Weipeng Xu, Kripasindhu Sarkar, and Christian Theobalt.
\newblock Estimating egocentric 3d human pose in global space.
\newblock In \emph{International Conference on Computer Vision (ICCV)}, 2021{\natexlab{a}}.

\bibitem[Wang et~al.(2022)Wang, Liu, Xu, Sarkar, Luvizon, and Theobalt]{wang2022estimating}
Jian Wang, Lingjie Liu, Weipeng Xu, Kripasindhu Sarkar, Diogo Luvizon, and Christian Theobalt.
\newblock Estimating egocentric 3d human pose in the wild with external weak supervision.
\newblock In \emph{Computer Vision and Pattern Recognition (CVPR)}, 2022.

\bibitem[Wang et~al.(2023)Wang, Luvizon, Xu, Liu, Sarkar, and Theobalt]{wang2023scene}
Jian Wang, Diogo Luvizon, Weipeng Xu, Lingjie Liu, Kripasindhu Sarkar, and Christian Theobalt.
\newblock Scene-aware egocentric 3d human pose estimation.
\newblock In \emph{Computer Vision and Pattern Recognition (CVPR)}, 2023.

\bibitem[Wang et~al.(2024)Wang, Cao, Luvizon, Liu, Sarkar, Tang, Beeler, and Theobalt]{wang2024egocentric}
Jian Wang, Zhe Cao, Diogo Luvizon, Lingjie Liu, Kripasindhu Sarkar, Danhang Tang, Thabo Beeler, and Christian Theobalt.
\newblock Egocentric whole-body motion capture with fisheyevit and diffusion-based motion refinement.
\newblock In \emph{Computer Vision and Pattern Recognition (CVPR)}, 2024.

\bibitem[Wang et~al.(2021{\natexlab{b}})Wang, Zhang, Cai, Yan, and Feng]{wang2021mvp}
Tao Wang, Jianfeng Zhang, Yujun Cai, Shuicheng Yan, and Jiashi Feng.
\newblock Direct multi-view multi-person 3d human pose estimation.
\newblock \emph{Advances in Neural Information Processing Systems (NeurIPS)}, 2021{\natexlab{b}}.

\bibitem[Xu et~al.(2019)Xu, Chatterjee, Zollhoefer, Rhodin, Fua, Seidel, and Theobalt]{xu2019mo2cap2}
Weipeng Xu, Avishek Chatterjee, Michael Zollhoefer, Helge Rhodin, Pascal Fua, Hans-Peter Seidel, and Christian Theobalt.
\newblock {Mo}$^{2}${Cap}$^{2}$ : Real-time mobile 3d motion capture with a cap-mounted fisheye camera.
\newblock \emph{IEEE Transactions on Visualization and Computer Graphics}, 2019.

\bibitem[Yang et~al.(2022)Yang, Guo, Zhang, and Wu]{yang2022u}
Honghong Yang, Longfei Guo, Yumei Zhang, and Xiaojun Wu.
\newblock U-shaped spatial-temporal transformer network for 3d human pose estimation.
\newblock \emph{Machine Vision and Applications}, 33\penalty0 (6):\penalty0 82, 2022.

\bibitem[You et~al.(2023)You, Liu, Wang, Li, Ding, and Li]{You_2023_ICCV}
Yingxuan You, Hong Liu, Ti Wang, Wenhao Li, Runwei Ding, and Xia Li.
\newblock Co-evolution of pose and mesh for 3d human body estimation from video.
\newblock In \emph{International Conference on Computer Vision (ICCV)}, 2023.

\bibitem[Yuan and Kitani(2019)]{yuan2019ego}
Ye Yuan and Kris Kitani.
\newblock Ego-pose estimation and forecasting as real-time pd control.
\newblock In \emph{International Conference on Computer Vision (ICCV)}, 2019.

\bibitem[Zhang et~al.(2022{\natexlab{a}})Zhang, Tu, Yang, Chen, and Yuan]{zhang2022mixste}
Jinlu Zhang, Zhigang Tu, Jianyu Yang, Yujin Chen, and Junsong Yuan.
\newblock Mixste: Seq2seq mixed spatio-temporal encoder for 3d human pose estimation in video.
\newblock In \emph{Computer Vision and Pattern Recognition (CVPR)}, 2022{\natexlab{a}}.

\bibitem[Zhang et~al.(2022{\natexlab{b}})Zhang, Ma, Zhang, Qian, Kwon, Pollefeys, Bogo, and Tang]{Zhang_ECCV_2022}
Siwei Zhang, Qianli Ma, Yan Zhang, Zhiyin Qian, Taein Kwon, Marc Pollefeys, Federica Bogo, and Siyu Tang.
\newblock Egobody: Human body shape and motion of interacting people from head-mounted devices.
\newblock In \emph{European conference on computer vision (ECCV)}, 2022{\natexlab{b}}.

\bibitem[Zhang et~al.(2023)Zhang, Ma, Zhang, Aliakbarian, Cosker, and Tang]{Zhang_2023_ICCV}
Siwei Zhang, Qianli Ma, Yan Zhang, Sadegh Aliakbarian, Darren Cosker, and Siyu Tang.
\newblock Probabilistic human mesh recovery in 3d scenes from egocentric views.
\newblock In \emph{International Conference on Computer Vision (ICCV)}, 2023.

\bibitem[Zhang et~al.(2021)Zhang, You, and Gevers]{zhang2021automatic}
Yahui Zhang, Shaodi You, and Theo Gevers.
\newblock Automatic calibration of the fisheye camera for egocentric 3d human pose estimation from a single image.
\newblock In \emph{Winter Conference on Applications of Computer Vision (WACV)}, 2021.

\bibitem[Zhao et~al.(2021)Zhao, Wei, Mahmud, and Frahm]{zhao2021egoglass}
Dongxu Zhao, Zhen Wei, Jisan Mahmud, and Jan-Michael Frahm.
\newblock Egoglass: Egocentric-view human pose estimation from an eyeglass frame.
\newblock In \emph{International Conference on 3D Vision (3DV)}, 2021.

\bibitem[Zhao et~al.(2023)Zhao, Zheng, Liu, Wang, and Chen]{Zhao_2023_CVPR}
Qitao Zhao, Ce Zheng, Mengyuan Liu, Pichao Wang, and Chen Chen.
\newblock Poseformerv2: Exploring frequency domain for efficient and robust 3d human pose estimation.
\newblock In \emph{Conference on Computer Vision and Pattern Recognition (CVPR)}, 2023.

\bibitem[Zheng et~al.(2021)Zheng, Zhu, Mendieta, Yang, Chen, and Ding]{Zheng_2021_ICCV}
Ce Zheng, Sijie Zhu, Matias Mendieta, Taojiannan Yang, Chen Chen, and Zhengming Ding.
\newblock 3d human pose estimation with spatial and temporal transformers.
\newblock In \emph{International Conference on Computer Vision (ICCV)}, 2021.

\bibitem[Zhou et~al.(2023)Zhou, Zhang, Hayder, Petersson, and Harandi]{Zhou_2023_ICCV}
Jieming Zhou, Tong Zhang, Zeeshan Hayder, Lars Petersson, and Mehrtash Harandi.
\newblock Diff3dhpe: A diffusion model for 3d human pose estimation.
\newblock In \emph{International Conference on Computer Vision (ICCV) Workshops}, 2023.

\bibitem[Zhu et~al.(2023)Zhu, Ma, Liu, Liu, Wu, and Wang]{Zhu_2023_ICCV}
Wentao Zhu, Xiaoxuan Ma, Zhaoyang Liu, Libin Liu, Wayne Wu, and Yizhou Wang.
\newblock Motionbert: A unified perspective on learning human motion representations.
\newblock In \emph{International Conference on Computer Vision (ICCV)}, 2023.

\bibitem[Zhu et~al.(2021)Zhu, Xu, Shen, Ji, Gao, and Shen]{zhu2021posegtac}
Yiran Zhu, Xing Xu, Fumin Shen, Yanli Ji, Lianli Gao, and Heng~Tao Shen.
\newblock Posegtac: Graph transformer encoder-decoder with atrous convolution for 3d human pose estimation.
\newblock In \emph{International Joint Conference on Artificial Intelligence (IJCAI)}, 2021.

\end{thebibliography}
}



\end{document}